\documentclass{article}

\usepackage[preprint]{neurips_2026}

\usepackage[utf8]{inputenc} %
\usepackage[T1]{fontenc}    %
\usepackage{hyperref}       %
\usepackage{url}            %
\usepackage{booktabs}       %
\usepackage{amsmath}        %
\usepackage{amsfonts}       %
\usepackage{nicefrac}       %
\usepackage{microtype}      %
\usepackage{xcolor}         %
\usepackage{xspace}         %
\usepackage{tikz}           %
\usepackage{siunitx}        %
\usepackage{multirow}       %
\usepackage{cleveref}       %
\usepackage{subcaption}     %
\usepackage{tabularx}       %
\usepackage{pifont}         %
\usepackage{xurl}           %

\newcommand{\sysname}{APWA\xspace}

\newcommand{\piitask}{\texttt{PII-300k}\xspace}
\newcommand{\sumtask}{\texttt{SummaryBench}\xspace}
\newcommand{\schematask}{\texttt{SchemaBench}\xspace}
\newcommand{\websurfer}{\texttt{WebSurfer}\xspace}

\newcommand{\mltables}{\textsc{MLTables}\xspace}
\newcommand{\discomat}{\textsc{DisCoMat}\xspace}
\newcommand{\swde}{\textsc{SWDE}\xspace}
\newcommand{\chemistry}{\textsc{Chemistry}\xspace}

\newcommand{\subrule}{\specialrule{0.2pt}{0.2em}{0.1em}}

\title{\sysname: A Distributed Architecture for Parallelizable Agentic Workflows}

\author{%
  Evan Rose \\
  Northeastern University\\
  \texttt{rose.ev@northeastern.edu} \\
  \And
  Tushin Mallick \\
  Northeastern University \\
  \texttt{mallick.tu@northeastern.edu} \\
  \And
  Matthew Laws \\
  Northeastern University \\
  \texttt{laws.ma@northeastern.edu} \\
  \And
  Cristina Nita-Rotaru \\
  Northeastern University \\
  \texttt{c.nitarotaru@northeastern.edu} \\
  \And
  Alina Oprea \\
  Northeastern University \\
  \texttt{a.oprea@northeastern.edu} \\
}

\begin{document}

\maketitle

\begin{abstract}
    Autonomous multi-agent systems based on large language models (LLMs) have demonstrated remarkable abilities in independently solving complex tasks in a wide breadth of application domains.
    However, these systems hit critical reasoning, coordination, and computational scaling bottlenecks as the size and complexity of their tasks grow. These limitations hinder multi-agent systems from achieving high-throughput processing for highly parallelizable tasks, despite the availability of parallel computing and reasoning primitives in the underlying LLMs.
    We introduce the \textit{Agent-Parallel Workload Architecture} (\sysname), a distributed multi-agent system architecture designed for the efficient processing of heavily parallelizable agentic workloads.
    \sysname facilitates parallel execution by decomposing workflows into non-interfering subproblems that can be processed using independent resources without cross-communication. It supports heterogeneous data and parallel processing patterns, and it accommodates tasks from a wide breadth of domains.
    In our evaluation, we demonstrate that \sysname can dynamically decompose complex queries into parallelizable workflows and scales on larger tasks in settings where prior systems fail completely.
\end{abstract}

\section{Introduction}
\begin{figure}[ht]
    \centering
    \includegraphics[width=1\linewidth]{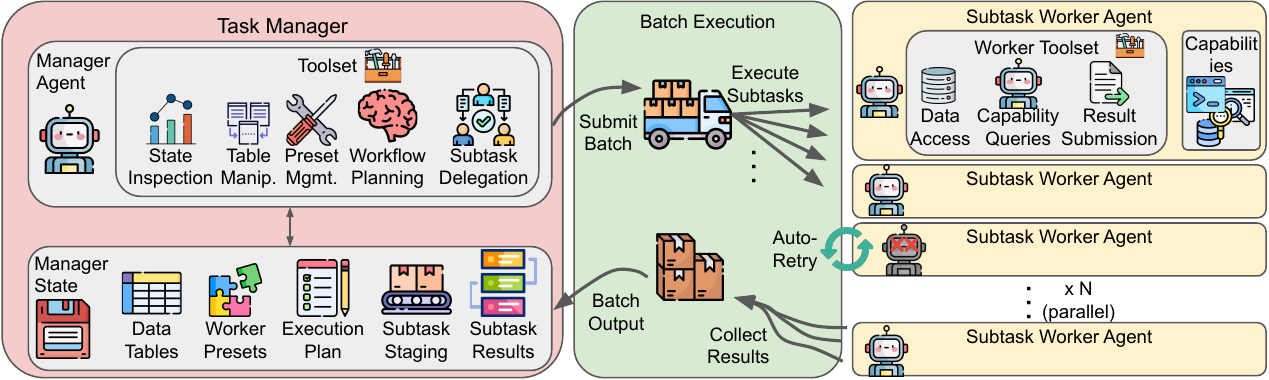}
    \caption{Overview of \sysname. \sysname dynamically decomposes tasks into parallelizable workflows leveraging agent workers and executes them in a distributed environment.}
    \label{fig:flow-overview}
\end{figure}

Autonomous LLM agents operationalize the planning, reasoning, and problem-solving capabilities of modern LLMs in real environments by situating the LLM at the center of a broader software system capable of interacting with an environment \cite{yao2023react,schick2023toolformer,shinn2023reflexion}.
These LLM agents have been successfully applied to domains spanning software development \cite{jimenez2024swebench,yang2024sweagent,wang2025openhands}, cybersecurity \cite{lee2024secbench,deng2024pentestgpt,meng2024large}, web browsing \cite{deng2023mind2web,zheng2024gpt4vision,qi2025webrl}, healthcare \cite{kim2024mdagents}, finance \cite{yu2024fincon}, scientific research \cite{boiko2023autonomous,bran2024augmenting,trirat2025automl,chan2025mlebench,yang2023leandojo}, and more.

While LLM agents have unlocked a wide array of applications, their LLM backbone imposes several technical constraints limiting their applicability in some settings.
While LLMs have absorbed knowledge from diverse domains (and thus are \textit{knowledge generalists}) \cite{brown2020language,hendrycks2021measuring}, they may struggle to reliably modulate between multiple functional roles within a single execution trajectory \cite{huang2024large,gupta2024llm}.
Additionally, as tasks become larger and more complex, reasoning and output quality degrade substantially \cite{liu2024lost,dziri2023faith,hsieh2024ruler}, and eventually collapse completely as the size of the data the LLM must process exceeds the context window of the LLM.
To scale agentic applications to a broader class of problems, research has explored developing \textit{multi-agent systems} which coordinate several LLM agents together via message passing to cohesively solve a single complex task \cite{hong2024metagpt,liang2024encouraging,qian2024chatdev,qian2025scaling,chen2024agentverse,fourney2024magentic-one,wang2025megaagent,hu2025owl}.
Multi-agent architectures allow agents to assume specialized roles \cite{li2023camel,qian2024chatdev,hong2024metagpt,wang2025megaagent,hu2025owl} and partition work, memory, and communication \cite{qian2025scaling}, leading to higher-quality outputs \cite{li2023camel,hong2024metagpt,wang2025mixture}, coherence over longer time scales \cite{hong2024metagpt,qian2025scaling}, and more efficient context management \cite{qian2025scaling,fourney2024magentic-one}.

Despite the progress made by prior research on multi-agent systems, 
existing solutions do not work well for problems that involve operating on a large amount of data that does not fit into an agent's context, memory, or permanent storage, or those involving a massive number of non-overlapping subproblems that each requires deeper research exploration. Specifically, they suffer from the following limitations. \textcircled{1} They do not provide support for intelligent and automated task decomposition over a distributed computing and storage infrastructure to support highly parallelizable workloads. \textcircled{2} They do not provide support for efficient and scalable coordination among large teams of agents. \textcircled{3} They limit an individual agent's exploration to solving a subproblem through excessive global coordination. \textcircled{4} They do not support complex, dynamic, data-dependent, heterogeneous planning and processing. \textcircled{5} They are not general enough, being designed for specific applications.

To better understand the problem we aim to solve, consider the paradigm shift introduced by systems designed for highly parallelizable workloads like MapReduce \cite{dean2004mapreduce} and Apache Spark \cite{zaharia2012resilient}. These systems transformed traditional data processing by simplifying the development of distributed applications. 
They provide intuitive and expressive abstractions for programmers, along with efficient and robust implementations of common data-processing patterns. As a result, they scale seamlessly in distributed environments while effectively hiding underlying system complexity from the user.

The research question that we are asking in this paper is how to create a similar paradigm shift in the AI agentic environment by designing an architecture that allows multi-agent systems to solve problems that would benefit from high-parallelization with respect to the  processed data and the executed subtasks.

Designing a system that meets our goals presents several challenges.
First, although LLMs have shown strong performance on software engineering tasks \cite{yang2024sweagent,jimenez2024swebench,wang2025openhands}, they lack native reasoning capabilities for distributed infrastructure and large-scale parallelism. Prior work on LLM-guided parallel workflows is limited \cite{kim2024llm,qiao2025benchmarking,niu2025flow} and does not address large-scale data processing.
Second, existing agent frameworks lack abstractions for reasoning over large, distributed datasets. LLM agents typically rely on local execution state, which does not scale when data exceeds a single machine or the model’s context window. Even metadata- or digest-based approaches break down at scale, as reasoning over millions of objects can overwhelm the model. Unlike traditional systems—where metadata is assumed to fit in working memory—LLM-based systems must contend with settings where even metadata is too large.

In this paper, we introduce the Agent-Parallel Workload Architecture (\sysname), a distributed multi-agent system designed specifically for processing parallelizable workflows. At the core of \sysname
lie novel programming abstractions suited for LLM agents that enable them to reason about large-scale data resources, decompose workflows into non-interfering subproblems that can be processed using independent resources without cross-communication, and issue and inspect distributed data-parallel execution flows. 
Based on these abstractions, we provide a scalable implementation of \sysname leveraging Ray \cite{moritz2018ray} and demonstrate that it accommodates different tasks that are parallelizable, outperforms several baselines, 
and scales on larger tasks in settings where prior systems fail completely.
We show an overview of \sysname in Figure \ref{fig:flow-overview}.

\section{Background and Problem Statement}
\subsection{Architectures for Multi-Agent Systems}
Several emerging architectures for multi-agent systems allow practitioners to express highly flexible multi-agent teams. However, they are not suitably designed to accommodate highly-distributed infrastructure, automated workload partitioning, and massively parallelized execution \cite{wu2024autogen,fourney2024magentic-one}.
Autogen \cite{wu2024autogen} implements a team-based programming model in which developers may manually define and instantiate agents within a publish-subscribe message sharing fabric.
Magentic-One \cite{fourney2024magentic-one}, a multi-agent system implemented in Autogen, features an orchestrator-worker architecture for automatically decomposing and solving complex tasks through multi-agent cooperation.
However, these agents operate via synchronized message passing explicitly routed through an LLM-based orchestrator agent, inhibiting scalability and parallelization since (1) the orchestrator LLM must manage a global view of the running agents, limiting the feasible number of agents to 10s or 100s, and (2) only a single agent can process information and communicate results at a time, making effective parallelization impossible.

A small set of multi-agent systems architectures attempt to facilitate parallelizable workflows.  However, they either restrict to static parallelization patterns \cite{langchain} or limit parallelization to a small scale by requiring expensive agent-modulated orchestration \cite{wang2025megaagent}.
For example, MegaAgent, a representative architecture in this category, has significant coordination overhead where administrators must {\em manually} create workers, can not accommodate format divergence between peers at different layers of the hierarchy, and can not organize large data across the workers.

\subsection{Problem Statement}
\label{sec:ps}
Our goal is to design a scalable multi-agent system suited for highly parallelizable agentic workflows meeting several design goals:

\textcircled{1} {\em Enable intelligent and automated task
decomposition over a distributing computing and storage infrastructure to support highly parallelizable workloads:} 
Our architecture will enable \textit{massively parallelized} execution patterns where many workflows can be partitioned into non-interfering subproblems that can be processed using independent resources and without cross-communication.
These parallel execution patterns are efficiently facilitated without resource or reasoning bottlenecks. 

\textcircled{2} {\em Provide support for efficient and scalable coordination among large teams of agents.}  Provide more efficient communication abstractions and state sharing including access to large data and a rich set of tools.

\textcircled{3} {\em Provide individual agents exploration capabilities for solving a subproblem.}  Our systems will allow
agents to perform individual exploration to solve a subproblem, by decoupling global coordination from local agent state.

\textcircled{4} {\em Provide support for complex, dynamic, data-dependent, heterogeneous planning and processing:} Our system will facilitate the execution of \textit{data-parallel} workflows (similar instructions applied to different subsets of a data resource), \textit{task-parallel} workflows (different instructions being applied to the same inputs), and \textit{replication-parallel} workflows (tasks and inputs are nearly identical but the goal of parallelization is to search through a large solution space more quickly).
Our system will facilitate data-dependent exploration for highly specialized task requests. 
In the case of heterogeneous data, simple data processing rules generated by a single source are insufficient to express the desired operations to apply to data.
Our system will support complex, context- and data-dependent trajectories which are hard to anticipate with static, prescriptive workflows. 

\textcircled{5} {\em Be task- and workflow agnostic:} Our system will automatically accommodate tasks from a wide breadth of domains and not restrict to particular processing or parallelization patterns. For example, our architecture \textit{naturally leverages} the expressivity and flexibility of LLM agents to automatically discover efficient parallel execution paths.

\paragraph{Challenges}
Designing a system meeting our design goals faces several key challenges.
First, while LLMs have made substantial progress in software engineering tasks \cite{yang2024sweagent,jimenez2024swebench,wang2025openhands}, they lack native reasoning capabilities for distributed infrastructure and large-scale parallelism.
Only very few works consider LLM-guided parallel workflow generation \cite{qiao2025benchmarking, niu2025flow}, and none of these address truly large-scale data processing.
Second, existing agent scaffolding does not provide the necessary abstractions for LLM-based components to reason over large and distributed data resources.
Typically, LLM agents operate by inspecting local execution state directly, but this approach does not scale when the data does not fit on a single machine, much less in the restricted context window of even modern LLMs.
In fact, even \textit{metadata-} or \textit{digest-based} approaches cannot work. In cases when the LLM must reason over thousands or even millions of data objects, even simply enumerating identifiers for each of those objects overloads the LLM's reasoning abilities.
Notice a key distinction from traditional computing settings: with traditional computing mechanisms (e.g., distributed dataflow engines), it is typically safe to assume that at least the \textit{metadata} fits inside the working memory of the system and is within its processing abilities.
With LLM-based components, however, \textit{even metadata becomes too large}.

\section{The \sysname System}
\label{sec:system-description}

\sysname is a distributed multi-agent system that is designed specifically for highly parallelizable agentic workflows. At the core of 
\sysname lies a set of abstractions that enables it to achieve all the five goals listed in \Cref{sec:ps}. These abstractions follow a hierarchical organization: system-level abstractions that support high-level parallelization goals (\Cref{sec:system-abstractions}), and distributed-level abstractions that support lower-level parallelization goals (\Cref{sec:system-dc-abstractions}). We use these abstractions to implement \sysname (Section \ref{sec:system-impl}).

\begin{figure}[ht]
    \centering
    \includegraphics[width=1\linewidth]{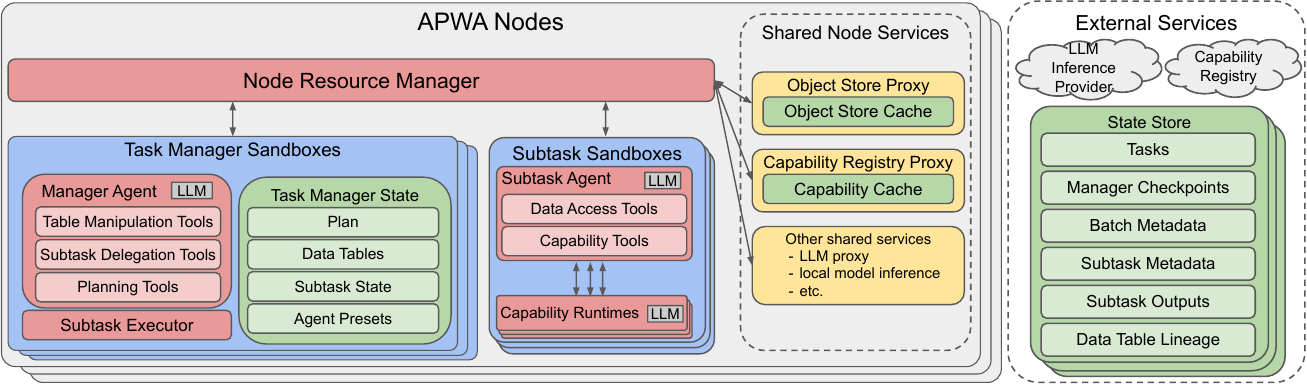}
    \caption{APWA Distributed System Architecture.}
    \label{fig:system-overview}
\end{figure}

\subsection{\sysname system-level abstractions}
\label{sec:system-abstractions}
As other multi-agent systems, \sysname is organized around three main abstractions: the \textit{manager}, the \textit{worker}, and the \textit{executor}.
However, there are fundamental differences with respect to other systems,
as these abstractions target data-parallel or task-parallel workflows. Both manager and worker have planning roles, the manager is performing meta-planning by decomposing the task into parallelizable subtasks, while a worker is performing planning to solve the assigned subtask, by using the necessary agents and tools. The executor performs the actual execution in a distributed cluster environment hiding the low-level distributed computing details.

\paragraph{Manager}
It is the abstraction in charge of solving a  task and there can be only one manager for a task. The manager dynamically \textit{partitions} the task into non-interfering \textit{subtasks} which can be dispatched and executed in parallel.
The manager is designed to perform parallelization by providing it with:
(1) definition of what is a general  parallelizable unit, i.e., subtask, (2) guided planning with examples of what parallelization means and  parallelization patterns, (3) abstractions for how large data can be communicated to workers. 
Subtasks specify contracts for a unit of work to be completed as well as the requested worker configuration.
The manager is the only component in the system with a global view of a task's execution state, and it is the only one performing high-level task reasoning.

\paragraph{Worker}
A worker represents the abstraction in charge of solving a subtask. There can be many workers (thousands) for a task. Workers can range in complexity from simple LLM processors to multi-agent subsystems with specialized roles and abilities. 
Workers also receive support for parallelization: 
(1) they receive guidance for subtask solving in terms of role they play in the system,
(2) they can use a wide set of agents and tools,
(3) they use abstractions to share large data across the agents they use to solve the task.
A worker has a \textit{local} view of the system state consisting of only  inputs explicitly passed by the manager and anything contained in their local execution space.

One of the key points of our design is identifying the appropriate separation between what is the global state maintained by the manager, and what is the local state maintained by the worker, as we wish to give workers high-autonomy on how they solve the subtask.

\paragraph{Executor}
The main role of the executor is to
collect execution requests from the manager, execute them and report the results. The fundamental difference from other systems, is that the focus is on leveraging distributed resources available in a cluster computing environment. Specifically, the executor places the subtasks on nodes in the cluster to execute, handling timeouts, automatic retries for transient errors, and principled collection of the results.
The executor abstracts away many of the distributing computing details, including worker placement across virtual or physical nodes, resource allocation, and failure recovery. By handling node-level resource management and automated retry logic internally, the executor allows the manager to focus entirely on the logical structure and semantics of the subtasks, without needing to reason about low-level infrastructure concerns.

\subsection{System abstractions for distributed agentic computation}
\label{sec:system-dc-abstractions}

To achieve our goal of a distributed system architecture for agentic parallelizable workflows, we introduce several key abstractions which enable the manager and workers to reason efficiently over intermediate execution state and distributed data resources, define and issue high volume of parallel subtasks, and  manipulate internal data representations.

\paragraph{Planning for parallelization}
We implement an instance of the planning design pattern tailored specifically for parallelizable workflows.
When the manager agent is initialized, it is instructed to immediately explore its local state in order to understand the nature of the task as well as the structure of any input data passed to it.
Then, the manager immediately generates a structured plan object consisting of a list of steps to be completed, that it will maintain for the duration of the task life cycle.
This plan includes fields that explicitly direct the agent to decompose the task into subtasks as well as discuss how to partition and/or reorganize any available data or intermediate results.
The manager is also permitted to specify an \textit{output contract} object, which defines output structures which must exist before the task can be considered completed and is checked against the true system state to provide feedback to the agent and prevent premature task termination.

\paragraph{Subtask delegation} The goal of this abstraction is to allow the manager to explain what a subtask is in a general manner without  manually generating a specification for each one. This abstraction is also designed to support parallelization by allowing
the manager to efficiently specify batches of \textit{subtasks} to be executed in parallel. We introduce \textit{subtask templates}, which describe a unit of work to be completed by a worker and include fields for agent configuration, the task to be completed, and references to data resources to be processed.
To enable efficient specification of a large number of subtasks, subtask template input parameters can contain placeholders which are automatically expanded when paired with an appropriate data resource, meaning the logical content of the subtask is decoupled from the scale of the data it acts upon.

\paragraph{Data tables}

In a distributed execution setting, some tasks might involve large data objects that collectively exceed the context window of the system LLM, or even the memory constraints of the machine processing the main task.
To minimize friction from large-scale data resources, we introduce a data table abstraction that facilitates easier data interactions with LLM-based components.
A data table is a logical, read-only, finite sequence of records conforming to a common \textit{schema}.
Data tables are quite similar to related notions in data analysis libraries and databases.
Importantly, data tables enable LLM-based components to interact with extremely large-scale data in a highly compact metadata representation.

\sysname's manager interacts with tables through a dedicated tool suite that enable the manager to inspect, query, analyze, manipulate, and construct tables dynamically in order to gather useful information about intermediate execution state as well as to prepare data for future processing. We define two categories of table-focused tools:
\textit{table analytics} tools allow the task manager agent to query information about data stored in tables and \textit{table manipulation} tools allow the task manager to construct new tables from existing ones.
A full set of table tools and their descriptions are given in \Cref{apx:task-manager-tools}.

\paragraph{Dynamic agent capabilities and presets}
Our system architecture is general-purpose, as we %
do not make  any explicit assumptions about which agent-powered software
capabilities are needed to perform a task.
Moreover, it may be the case that these capabilities must be determined dynamically at runtime, perhaps as part of the task specification itself.
One implication of this fact is that our system should allow the definition, construction, and execution of specialized agents at runtime.

We realize our goal of dynamic agents at runtime with an \textit{agent capability and preset} abstraction. This provides a useful abstraction for the manager to reason about how to complete subtasks correctly despite remaining a general-purpose component. Certain tasks may require specialized tools or processing steps which the manager can identify and leverage.

Runtime capabilities are discovered through an externally hosted \textit{Capability Registry} which exposes diverse software functionalities (\textit{capabilities}) 
that can be used during subtask execution.
These capabilities enable dynamic, composable functionalities for workers.
To create agents using capabilities from the registry, the task manager agent creates an \textit{agent preset} that defines a system prompt and a set of capabilities for the worker agent to use. When delegating subtasks, the manager agent specifies which preset to use.

\subsection{\sysname implementation}
\label{sec:system-impl}
Figure \ref{fig:system-overview} depicts the architecture of our system
using the abstractions described above. 
At runtime, \sysname processes  a series of \textit{rounds} cycling between two phases: a high-level reasoning and delegation phase performed by the manager and a low-level subtask execution phase performed by workers. Both the manager and the worker are implemented as agents running in sandboxed  environments.

\paragraph{Manager environment}
The manager's core responsibility is to oversee the successful completion of a single task. The manager does this by planning out appropriate processing steps, observing and manipulating local data representations, and dispatching units of work to be executed remotely.
The manager keeps track of progress of each subtask, by maintaining  a highly-compact metadata representation of the global system data, a collection of workers, an actively-maintained execution plan, a detailed output format specification, and a log of results from previous execution rounds.

The manager also maintains its internal execution state in an external state store that records task metadata, manager checkpoints, issued subtasks, subtask outcomes, and synthesized results. This store serves as the primary interface through which the manager interacts with system state, allowing it to retrieve outputs from completed subtask run and use them in analysis or re-planning. Each subtask run contributes structured updates to this state, including status indicators, metrics, and artifact references.

\paragraph{Worker environment}
The worker's execution environment is equipped with whatever auxiliary runtime capabilities are needed to complete the task, such as virtual machine resources (local file system, terminal capabilities, etc.) or network-accessed services. The worker has tools for reading and writing data to a shared network-accessible data resource.

We observe that frequently agent execution patterns reduce to a simple LLM invocation. For example, simple data processing steps like summarization or structured information extraction involve only a single LLM generation step. In these cases, we can improve subtask throughput by a substantial factor (10-100$\times$) by dynamically routing the subtask execution to a lightweight execution space.
This routing is determined dynamically by the manager when it emits a subtask based on its own assessment of the complexity of the subtask.

\paragraph{Scalable fabric for executor}
We use the Ray distributed computing framework \cite{moritz2018ray} to provide \sysname's distributed computing fabric.
Ray provides a task-parallel programming model and system implementation that scales efficiently to large-scale distributed computing resources.
Ray features a distributed scheduler and can scale to hundreds of thousands of concurrent tasks and a throughput of millions of tasks per second. Multiple workers can be deployed on a given node.

When scaling to large numbers of subtasks, the probability that at least one failure occurs increases dramatically.
Many of these errors are transient (e.g., a temporary network partition) and not critical errors that must be resolved at the logical layer.
Our execution layer automatically retries subtasks up to a fixed number of times before returning a logical failure to the task manager.
This behavior allows the task manager to more reliably reason about the \textit{logical} content of its subtasks rather than low-level behaviors arising from the unreliable distributed system.

\section{Evaluation}
\label{sec:evaluation}

\subsection{Main experiments}
\label{sec:evaluation-main}

\paragraph{Benchmarks}
We evaluate \sysname on two public benchmarks, the AI4Privacy PII-300k dataset (\piitask) \cite{ai4privacy2024pii} and a schema-driven structured content extraction benchmark (\schematask) \cite{bai2024schema}, as well as a manually-collected hierarchical summarization benchmark (\sumtask). We present summarized results here and provide full results, with variance measurements, in \Cref{apx:extended-experiments}.

The goal of PII-300k  is to redact sensitive fields in a large number of unstructured records, with labels from 27 PII categories targeting domains spanning education, health, and psychology.
The goal of SchemaBench is to extract valid and correct JSON outputs given a collection of documents from heterogeneous domains and data formats, including \LaTeX, XML, CSV, and HTML.
These first two tasks represent highly data-parallelizable processing tasks, as might be performed with standard distributed data-flow execution engines.
In both cases, ML-based methods are preferable to heuristic rule-based approaches due to the rich semantic nature of the task.

The goal of hierarchical summarization is to process a hierarchically-organized data corpus and generate summaries at different levels of granularity matching the granularity of the original data's organization pattern.
For hierarchical summarization, we collect three literary corpora of varying sizes: \textit{Romeo and Juliet} \cite{shakespeare1998romeo}, \textit{The Dynasts} \cite{hardy2005dynasts}, and \textit{The History of the Decline and Fall of the Roman Empire} \cite{gibbon2008decline}.
We obtain the source texts in plain text format from Project Gutenberg \cite{gutenberg} and manually parse and format each one into an appropriate organization pattern.
The data we use for \sumtask spans from small (2 tiers, \qty{166}{\kilo\byte}) to moderate (3 tiers, \qty{942}{\kilo\byte}) to large (4 tiers, \qty{10.5}{\mega\byte}), with statistics on each tier provided in \Cref{tab:summarization-data}.
The hierarchical summarization task represents a more complex parallelization topology which requires issuing multiple rounds of subtasks to be completed in parallel and combining results from previous rounds.

Together, these benchmarks measure the ability of \sysname to automatically decompose a complex task a parallelizable workflow, execute that workflow, reason over progress and results, and synthesize a comprehensive answer.

\paragraph{Metrics}
For each benchmark, we define an automated evaluation procedure to score the effectiveness of \sysname along the dimensions of \textit{utility} (whether the system produces a correct output) and \textit{cost} (wall-clock runtime, token usage, and monetary expense).
In general, we decompose utility into two submetrics, \textit{structural} (whether the outputs exist and are correctly formatted) and \textit{semantic} (whether the outputs are correct).
The precise definition of these submetrics is task dependent; due to space constraints we give brief descriptions here and defer detailed descriptions to \Cref{apx:extended-experiments}.

For \piitask, the structural score measures the correctness of the output table schema and number of emitted entries, while the semantic score measures the macro-averaged $F_1$ score of detected PII instances across all PII categories.
For \schematask, the structural score is similar to \piitask, while the semantic score reports the same metrics per extraction task as the original work \cite{bai2024schema}, typically a micro- or macro-$F_1$ score.
For \sumtask, we report the mean layer-wise ROUGE1-$F_1$ score averaged across trials against a set of reference summaries.

\paragraph{Baselines}
For every benchmark, we consider three baselines.
First, we process the entire task request, including any necessary data formatted in plain text, directly with an LLM.
In this scenario, we request the LLM to generate its response subject to a task-specific schema using the structured generation modes supported by recent LLM inference engines.

Second, we use the Magentic-One multi-agent system \cite{fourney2024magentic-one} to process the query.
Magentic-One represents a state-of-the-art multi-agent scaffold with built-in file system, web browser, and coding agent capabilities, as well as a dedicated planning and agent routing module.
Magentic-One executes worker agents serially with coordination handled by a dedicated orchestrator agent.
Because tasks are completed by interacting with tools and data file system, Magentic-One is better able to process larger data scales than na\"ively submitting the full file contents to LLM inference.

Third, we compare with the MegaAgent \cite{wang2025megaagent} multi-agent system.
MegaAgent is designed as a general-purpose multi-agent architecture with considerations for parallel agent execution.
As a multi-agent system, MegaAgent is logically organized into a hierarchical coordination structure.
A single \textit{boss} agent initiates task execution, and can spawn teams of subagents to help partition work.
This process continues recursively with privileged subagents themselves spawning subteams until a subtask is complete or a recursion limit is reached.

\paragraph{System configuration}
For our main experiments, we run \sysname on a single machine running Ubuntu 22.04.5 and equipped with an AMD Ryzen Threadripper PRO 5955WX 16-Core processor, \qty{252}{\giga \byte} memory, and 2x NVIDIA GeForce RTX 4090 GPUs.
We experiment using different combinations of GPT-5.4, GPT-5.4-mini, and GPT-5.4-nano \cite{openai2025gpt54,openai2025gpt54mini} for the manager agent and worker  LLM backends.
We use the same models to evaluate the baseline methods.

\begin{table*}[t]
\centering
\scriptsize
\setlength{\tabcolsep}{2pt}

\caption{Failure rates and wall-clock runtimes for \sumtask{} and \piitask{}. \sysname leverages parallelization to efficiently solve tasks with increasingly large input sizes. All settings use GPT-5.4 mini, except MegaAgent using GPT-4.1 mini. Wall-clock runtimes reflected only for trials that produced output. FR = failure rate, WC = wall-clock time (seconds).}

\begin{tabular}{l cc cc cc cc cc cc}
\toprule
& \multicolumn{2}{c}{R\&J} & \multicolumn{2}{c}{Dynasts} & \multicolumn{2}{c}{Roman}
& \multicolumn{2}{c}{PII-64} & \multicolumn{2}{c}{PII-512} & \multicolumn{2}{c}{PII-4096} \\
\cmidrule(lr){2-3}\cmidrule(lr){4-5}\cmidrule(lr){6-7}
\cmidrule(lr){8-9}\cmidrule(lr){10-11}\cmidrule(lr){12-13}
\textbf{Method}
& \textbf{FR\,($\downarrow$)} & \textbf{WC\,(s)\,($\downarrow$)}
& \textbf{FR\,($\downarrow$)} & \textbf{WC\,(s)\,($\downarrow$)}
& \textbf{FR\,($\downarrow$)} & \textbf{WC\,(s)\,($\downarrow$)}
& \textbf{FR\,($\downarrow$)} & \textbf{WC\,(s)\,($\downarrow$)}
& \textbf{FR\,($\downarrow$)} & \textbf{WC\,(s)\,($\downarrow$)}
& \textbf{FR\,($\downarrow$)} & \textbf{WC\,(s)\,($\downarrow$)} \\
\midrule
Direct & 0\% & 19 & 60\% & 76 & $100\%$ & $\bot$ & 0\% & 37 & 0\% & 41 & $100\%$ & $\bot$ \\
Magentic-One & $100\%$ & $\bot$ & $100\%$ & $\bot$ & $100\%$ & $\bot$ & $100\%$ & $\bot$ & $100\%$ & $\bot$ & 80\% & 91 \\
MegaAgent & 80\% & 472 & 80\% & 248 & 70\% & 579 & 60\% & 390 & 80\% & 25 & 70\% & 372 \\
\midrule
\sysname & 0\% & 157 & 0\% & 210 & 0\% & 329 & 0\% & 35 & 0\% & 67 & 0\% & 221 \\
\bottomrule
\end{tabular}
\label{tab:main-frwc-results}
\end{table*}

\begin{table*}[t]
\centering
\scriptsize
\setlength{\tabcolsep}{3pt}

\caption{Structural and semantic scores for \sumtask{} and \piitask{}. \sysname preserves structural and semantic integrity with increasingly large input sizes. All  settings  use GPT-5.4 mini, except MegaAgent using GPT-4.1 mini. Scores computed only for trials that produced output. Str. = structural score, Sem. = semantic score.}

\begin{tabular}{l cc cc cc cc cc cc}
\toprule
& \multicolumn{2}{c}{R\&J} & \multicolumn{2}{c}{Dynasts} & \multicolumn{2}{c}{Roman}
& \multicolumn{2}{c}{PII-64} & \multicolumn{2}{c}{PII-512} & \multicolumn{2}{c}{PII-4096} \\
\cmidrule(lr){2-3}\cmidrule(lr){4-5}\cmidrule(lr){6-7}
\cmidrule(lr){8-9}\cmidrule(lr){10-11}\cmidrule(lr){12-13}
\textbf{Method}
& \textbf{Str.\,($\uparrow$)} & \textbf{Sem.\,($\uparrow$)}
& \textbf{Str.\,($\uparrow$)} & \textbf{Sem.\,($\uparrow$)}
& \textbf{Str.\,($\uparrow$)} & \textbf{Sem.\,($\uparrow$)}
& \textbf{Str.\,($\uparrow$)} & \textbf{Sem.\,($\uparrow$)}
& \textbf{Str.\,($\uparrow$)} & \textbf{Sem.\,($\uparrow$)}
& \textbf{Str.\,($\uparrow$)} & \textbf{Sem.\,($\uparrow$)} \\
\midrule
Direct & 1.000 & 0.433 & 0.979 & 0.210 & $\bot$ & $\bot$ & 1.000 & 0.775 & 0.750 & 0.162 & $\bot$ & $\bot$ \\
Magentic-One & $\bot$ & $\bot$ & $\bot$ & $\bot$ & $\bot$ & $\bot$ & $\bot$ & $\bot$ & $\bot$ & $\bot$ & 1.000 & 0.179 \\
MegaAgent & 0.140 & 0.043 & 0.212 & 0.023 & 0.160 & 0.016 & 0.375 & 0.000 & 0.000 & 0.000 & 0.250 & 0.000 \\
\midrule
\sysname & 0.954 & 0.424 & 0.954 & 0.419 & 0.919 & 0.232 & 1.000 & 0.759 & 1.000 & 0.772 & 0.900 & 0.544 \\
\bottomrule
\end{tabular}
\label{tab:main-strsem-results}
\end{table*}

\paragraph{Results}
Table~\ref{tab:main-frwc-results} shows task completion rates and wall clock runtimes for \sumtask and \piitask, and \Cref{tab:main-strsem-results} reports output quality scores for \sumtask and \piitask.
We find that failure rates increase and output quality decreases for all baselines as the number of processed documents increases. Direct LLM use solves the task for small inputs (on \textit{Romeo and Juliet} and \piitask up to 512 records), but fails on larger inputs (on \textit{Roman} and \piitask with 4096 records).  
Magentic-One is unable to complete any of the tasks reliably, only managing to produce an output in a single trial for \piitask with all 4096 documents.
We identify two failure modes in Magentic-One: \textit{orchestration failure} and \textit{context explosion}.
In general, Magentic-One attempted to solve the task using programmatic methods rather than raw LLM processing.
Orchestration failure occurred when the Magentic-One orchestrator incorrectly routed requests between its coding and terminal agents, causing both subagents to fail to complete any useful work.
Context explosion occurred on larger inputs: for the \textit{Roman} corpus, Magentic-One's FileSurfer agent tried to page through the entire list of over 2,500 input documents and consumed over 6.5M input tokens, blocking progress until a timeout at 30 minutes.
MegaAgent experienced different scaling and coordination difficulties.
In one summarization task run, MegaAgent proceeds by creating a primary task orchestrator agent, which partitions work between three subagents (``SceneSummarizer'', ``ActSummarizer'', and ``FullSummarizer'').
The SceneSummarizer fails to finish processing the scene files quickly, and the orchestrator prematurely terminates the task.
In other runs, the agents produce large output files with incorrectly formatted JSON, causing the agent harness to crash.

In contrast, \sysname scales naturally to larger and more complex corpora, preserving better output fidelity as well as completing large tasks in a reasonable amount of time.
Despite the size of input data increasing by nearly two orders of magnitude (\qty{166}{\kilo\byte} on \textit{Romeo and Juliet} $\rightarrow$ \qty{10500}{\kilo\byte} on \textit{Roman}), \sysname's wall-clock runtime only increases moderately (\qty{157}{\second} $\rightarrow$ \qty{329}{\second}) thanks to massive parallelization, running over 2.5k agents concurrently.
Prior methods fail to generate results at all.
Table \ref{tab:main-sysname-ablation} shows that \sysname with different model configurations is able to process increasingly more complex tasks--growing in both raw input size and number of required processing steps--while effectively leveraging parallelization to keep runtime low. The use of larger models during planning (GPT-5.4) provides an advantage on the system's utility, achieving higher structural and semantic scores, at marginal cost increase.

Our qualitative findings on \schematask given in \Cref{apx:full-schema-results} are similar to our findings on \sumtask and \piitask for both runtime and output quality.
On the only dataset whose size is small enough to fit within the 1M token limit for GPT-5.4 mini, \chemistry, the semantic scores for \sysname are comparable to the scores for querying the LLM directly.
For all other datasets, the input size exceeds the context window of the considered LLMs when formatted directly.
Magentic-One encounters similar orchestration failures, and only manages to produce output in 4 out of 20 cases.
Like in \sumtask and \piitask, Magentic-One pursues programmatic solutions that do not meet the complex semantic nature of the task.
MegaAgent also fails to complete the structured extraction task, again due to a failure to parallelize work and instead attempt purely role-based task decomposition.
Out of 20 attempted tasks, only 6 produced any output, only 3 of which received nonzero scores. Of these, the semantic score is always 0.

\begin{table*}[t]
\centering
\scriptsize
\setlength{\tabcolsep}{1.5pt}

\caption{Comparison of \sysname{} configurations on \sumtask{}. Str. = structural score, Sem. = semantic score, WT = wall-clock time (sec), Cost = total cost.}
\label{tab:main-sysname-ablation}

\begin{tabular}{l cccc cccc cccc}
\toprule
& \multicolumn{4}{c}{R\&J} 
& \multicolumn{4}{c}{Dynasts} 
& \multicolumn{4}{c}{Roman} \\
\cmidrule(lr){2-5}\cmidrule(lr){6-9}\cmidrule(lr){10-13}
\textbf{Method}
& \textbf{Str.\,($\uparrow$)} & \textbf{Sem.\,($\uparrow$)} & \textbf{WT\,(s)\,($\downarrow$)} & \textbf{Cost\,($\downarrow$)}
& \textbf{Str.\,($\uparrow$)} & \textbf{Sem.\,($\uparrow$)} & \textbf{WT\,(s)\,($\downarrow$)} & \textbf{Cost\,($\downarrow$)}
& \textbf{Str.\,($\uparrow$)} & \textbf{Sem.\,($\uparrow$)} & \textbf{WT\,(s)\,($\downarrow$)} & \textbf{Cost\,($\downarrow$)} \\
\midrule
\sysname (5.4 $\times$ mini) & 1.000 & 0.528 & 152 & \$0.628 & 0.997 & 0.451 & 212 & \$1.16 & 0.983 & 0.370 & 336 & \$6.57 \\
\sysname (5.4 $\times$ nano) & 0.943 & 0.439 & 157 & \$0.582 & 0.923 & 0.419 & 211 & \$0.919 & 0.872 & 0.395 & 352 & \$3.09 \\
\sysname (mini $\times$ mini) & 0.897 & 0.394 & 94 & \$0.294 & 0.954 & 0.419 & 134 & \$0.728 & 0.817 & 0.207 & 235 & \$5.24 \\
\sysname (mini $\times$ nano) & 0.916 & 0.408 & 96 & \$0.225 & 0.803 & 0.314 & 149 & \$0.404 & 0.853 & 0.283 & 265 & \$2.08 \\
\bottomrule
\end{tabular}
\end{table*}

\subsection{Web browsing experiment}
\label{sec:evalation-web}

To demonstrate the utility of \sysname in orchestrating fully agentic workloads, we consider a technical report synthesis task leveraging web browsing agents.
The task definition is to generate textual reports on a list of topics provided in the input.
We consider two variants with a small (10), medium (20), and large (100) number of topics, respectively.
We populate the capability registry with Magentic-One's web surfer agent and expose its description to the manager agent.
We measure whether the correct number of (topic, report) pairs are generated and use LLM-as-a-judge to measure whether the report is factual and on-topic.
In experiments, \sysname successfully identifies the web surfer agent as relevant and creates an appropriate subtask template to pass input topics to workers. \sysname always achieves structural and semantic scores above 0.975.
\sysname incurred an execution time of \qty{143}{\second}, \qty{157}{\second}, and \qty{595}{\second} for 10, 20, and 100 topics, respectively. 
These results demonstrate \sysname's ability to leverage agents within a parallelizable workflow to complete a complex task efficiently.
Full results are given in \Cref{apx:full-websurfer-results}.

\section{Conclusion}
\label{sec:conclusion}

In this work, we propose \sysname, a highly scalable multi-agent execution architecture for parallelizable agentic workflows.
\sysname introduces efficient abstractions for LLM-based agents to reason about and act upon high volumes of distributed data and computing resources and to specify and issue large-scale parallel execution of agents.
We demonstrate that \sysname is capable of automatically and efficiently solving large-scale distributed workflows, in contrast to existing methods which cannot effectively leverage parallelism and fail on larger complex tasks.

\begin{ack}

This research was supported by funding from Google and Good Ventures foundation.

\end{ack}

\bibliographystyle{plain}
\bibliography{references}

\appendix
\crefalias{section}{appendix}
\crefalias{subsection}{appendix}
\crefalias{subsubsection}{appendix}

\section{Limitations}
\label{apx:limitations}

While \sysname makes important progress towards parallelized agentic task processing, we recognize some of its limitations. First, the architecture does not support direct communication between workers as workers interact exclusively with managers, which limits applicability to workflows requiring coordination among subtasks. 

Second, APWA's task planning provides guidance for using several  parallelization patterns and templates. However, we have not evaluated whether APWA generalizes naturally to other patterns. 

Third, security and privacy concerns were not considered in the current design. APWA's increased autonomy and throughput may  lead to vulnerabilities from prompt injection attacks, malicious tools, and compromised workers, as well as risks of data leakage. Future work should systematically analyze APWA's attack surface and develop guardrails for secure deployment.

\section{Broader impacts}
\label{apx:societal-impacts}

APWA enables multi-agent systems to efficiently process complex, large-scale tasks. By enabling dynamic workflow decomposition and parallel execution across heterogeneous data and processing patterns, APWA can accelerate  applications in different domains where current multi-agent systems fail to scale. We envision positive impacts to society if APWA is deployed in high-stake domains such as health care that require large-scale data processing. 

As with any system that increases the autonomy and scale of LLM-based agents, APWA carries potential misuse risks. The same parallelization capabilities that enable legitimate high-throughput workflows could be exploited to automate harmful activities at scale, such as  coordinated cyber attacks, large-scale phishing, or spam generation. Mitigating these threats requires future research on deploying safety guardrails, as well as defining and enforcing policies on the system's use.

\section{Related work}
\label{sec:related-work}

\textit{Distributed execution architectures} \cite{dean2004mapreduce,hadoop,isard2007dryad,power2010piccolo,malewicz2010pregel,murray2011ciel,zaharia2012resilient,rocklin2015dask,moritz2018ray} have studied ways to leverage distributed computing infrastructure for the efficient execution of data- and task-parallel workflows. Notably, MapReduce\cite{dean2004mapreduce} introduces a programming model and processing engine capable of handling massive datasets in parallel across distributed, commodity hardware. The innovation of MapReduce spurred heightened research into distributed execution architectures that build on top of the idea and improve on it by using abstractions to model the workflows \cite{hadoop,zaharia2012resilient,moritz2018ray}.

LLM-based multi-agent systems are evolving along several threads. General purpose orchestration substrates expose conversational, role-based or graph-based abstractions to compose co-operating agents \cite{wu2024autogen,hong2024metagpt,li2023camel,zhuge2024gptswarm}, a line increasingly mirrored by deployed industry frameworks and systems \cite{crewai,openaiagents,msagents,fourney2024magentic-one}. Building on these substrates, role-specialized teams emulate human workflows in domains such as SOP-driven software engineering~\cite{qian2024chatdev,dong2024selfcollaboration}. A separate thread improves reasoning through debate and consensus \cite{du2024improving,liang2024encouraging}, while others push towards scalability by dynamically generating agents without predefined SOPs \cite{wang2025megaagent,chen2024autoagents,hu2025owl}.

\section{\sysname implementation details}
\label{apx:system-details}
In this appendix, we give a detailed description of all system components, core distributed computing abstractions, and external resources used by \sysname and introduced in \Cref{sec:system-description}.
First, we cover the core system-level abstractions presented in \Cref{sec:system-abstractions}: the \textit{Task Manager} (\Cref{apx:system-task-manager}), the \textit{Worker} (\Cref{apx:system-worker}), and the \textit{Executor} (\Cref{apx:system-executor}).
Then, we provide a detailed description of the agentic abstractions from \Cref{sec:system-dc-abstractions}: the \textit{parallel planning} abstraction (\Cref{apx:system-parallel-planning}), the \textit{subtask delegation} abstraction (\Cref{apx:system-subtask-delegation}), the \textit{data table} abstraction (\Cref{apx:system-tables}), and the \textit{dynamic agent capability and preset} abstraction (\Cref{apx:system-capability-registry}).
Finally, we discuss practical implementation details for realizing \sysname using scalable infrastructure, including data storage services (\Cref{apx:system-data}), and considerations for computing cluster resource management (\Cref{apx:system-cluster}).

\subsection{Task manager}
\label{apx:system-task-manager}

The \textit{Task Manager} (or just \textit{Manager}) is the system component responsible for overseeing the life cycle of a single user task.
It receives a task from the orchestrator enclosing the user query expressed in natural language and any pre-configured environment state in the form of references to external data located in the object store.

\subsubsection{Manager agent}
The core operative backbone of the Manager is its namesake \textit{manager agent}.
The logical role of the manager agent is conceptually similar to other hierarchically-organized multi-agent system architectures like Magentic-One \cite{fourney2024magentic-one} and MegaAgent \cite{wang2025megaagent}, in that the manager agent is responsible for high-level planning, reasoning over intermediate global state, and coordinating subordinate subagents in order to make progress towards task completion.
In addition to these roles, we also design the manager agent to be tightly integrated with our core system abstractions.
These abstractions facilitate agent-based reasoning for parallel processing workflows over distributed computing resources.

\paragraph{Agent organization}
The per-round exploration phase, performed by the manager agent, involves three distinct steps.
The first, most substantial step, is the iterative tool-calling loop which the agent uses to inspect and interact with the environment.
During this step, the manager agent plans, reads execution state, creates agent presets, manipulates data tables, and stages subtasks to make progress towards task processing.
Following this step, a non-tool-using \textit{report generation} step is performed to extract a structured object describing any accomplishments, next steps, or critical blockers from the round.
This intermediate step is primarily for observability and debugging purposes.
Finally, a \textit{delegation} step validates the coherence of the current agent trajectory against the task, plan, and output contract and determines if the proposed system transition (either subtask batch emission or task finalization) is appropriate.
In the case that the transition is inappropriate, control is returned to the first step.
This final step is motivated by the agent debate design pattern \cite{du2024improving,liang2024encouraging}, and is intended to help detect and resolve erroneous or premature state transitions.

\paragraph{LLM context formation}
To integrate our core abstractions with the manager agent, we must organize and format system state information into reasonably succinct textual representations to be fed into LLM-based components.
Following existing agentic systems, we organize the system context as a chat-like transcript.

Starting with the manager agent's system prompt, we aim to provide a sufficiently coherent system model and catalog of domain concepts to the LLM.
In particular, we describe the \sysname system and the manager agent's operative role within it in the system prompt.
Additionally, we describe all domain concepts, including the core \sysname abstractions.
Then, to better condition the model towards generating coherent parallelizable workflows, we provide several examples of generic parallel computation patterns and how to realize them.

\subsubsection{Manager agent tools}
\label{apx:task-manager-tools}
One of the crucial abstractions implemented in the task manager is the \textit{comprehensive tool suite} designed to facilitate agent-based reasoning over distributed execution state.
The tool suite conceptually consolidates all of our distributed agentic computation abstractions into a single unified programming interface connected to the Manager Agent.
This tool suite extends the functionality of the base LLM agent to enable it to reason over parallelization patterns and inspect intermediate state in a way that carefully balances exploration with context efficiency.
\Cref{tab:agent-tools} lists all tools made available to the manager agent.

\begin{table}[htb]
\centering
\footnotesize
\caption{Manager agent tools. These tools serve as the interface between the LLM and the key system abstractions discussed in \Cref{sec:system-dc-abstractions}, allowing the manager agent to inspect system state, manipulate data, and delegate subtasks.}
\begin{tabularx}{\linewidth}{@{}>{\raggedright\arraybackslash}p{0.27\linewidth}X@{}}
\toprule
Tool & Description \\
\midrule
\texttt{list\_tables} & Lists available tables, grouped by table kind. \\
\texttt{get\_table\_meta} & Returns metadata for a table, either as a full schema/lineage view or a compact row-and-column summary. \\
\texttt{preview\_rows} & Shows a paginated preview of rows from a table, optionally selecting columns and ordering results. \\
\texttt{get\_row} & Retrieves one row from a table by row ID, optionally limited to selected columns. \\
\texttt{filter\_rows} & Returns rows from a table that match a structured predicate. \\
\subrule
\texttt{distinct\_values} & Lists distinct values for a field, along with counts, optionally under a predicate filter. \\
\texttt{value\_counts} & Computes the top-$k$ most frequent values for a field, with an optional predicate filter. \\
\texttt{summarize\_numeric} & Computes summary statistics for one or more numeric fields, optionally over filtered rows. \\
\texttt{groupby\_aggregate} & Performs grouped aggregation with count, count-distinct, sum, average, min, or max aggregations. \\
\texttt{sample\_rows} & Draws a random sample of rows from a table, optionally selecting columns and using a seed. \\
\subrule
\texttt{create\_union\_table} & Creates a derived table by vertically concatenating multiple schema-compatible tables. \\
\texttt{create\_filtered\_table} & Creates a derived table containing only rows from an input table that satisfy a predicate. \\
\texttt{create\_projected\_table} & Creates a derived table containing only a specified subset of columns. \\
\texttt{create\_joined\_table} & Creates a derived table by joining two input tables on specified key pairs. \\
\texttt{create\_grouped\_table} & Creates a derived table by grouping rows and applying named first, count, or collect aggregations. \\
\texttt{rename\_table} & Renames a table’s display name without changing its data or lineage. \\
\texttt{archive\_table} & Archives a table so it is hidden from ordinary table listings while preserving lineage. \\
\texttt{unarchive\_table} & Restores an archived table to the active namespace, optionally assigning a display name. \\
\texttt{rename\_columns} & Creates a derived table with selected columns renamed. \\
\texttt{add\_computed\_columns} & Creates a derived table with new computed columns using cast, concat, coalesce, key extraction, or first-element extraction. \\
\texttt{create\_results\_with\_source} & Joins a batch results table back to its discovered source table using lineage columns. \\
\texttt{drop\_columns} & Creates a derived table with specified columns removed. \\
\subrule
\texttt{stage\_single\_subtask} & Adds one literal subtask definition to the staging area. \\
\texttt{stage\_dataset\_subtask} & Stages a batch of subtasks by applying a template to every row of a dataset table. \\
\texttt{remove\_staged\_subtask} & Removes one staged subtask template. \\
\texttt{clear\_staged\_subtasks} & Clears all currently staged subtasks from the staging area. \\
\texttt{list\_subtasks} & Lists executed subtasks with pagination and an optional status filter. \\
\texttt{get\_subtask\_result} & Fetches detailed run results, metrics, status, and artifacts for a specific subtask ID. \\
\texttt{get\_artifact} & Retrieves artifact metadata and, optionally, truncated artifact content. \\
\texttt{list\_artifacts} & Lists artifacts for the task, optionally filtered by subtask and optionally including previews. \\
\subrule
\texttt{write\_plan} & Replaces the manager plan with a new full plan payload. \\
\texttt{write\_output\_contract} & Replaces the task output contract and reports whether the current dataset satisfies it. \\
\subrule
\texttt{create\_agent\_preset} & Creates or updates an agent preset with a prompt and capability list. \\
\texttt{list\_agent\_presets} & Lists all currently available agent presets. \\
\texttt{delete\_agent\_preset} & Deletes an agent preset by name. \\
\bottomrule
\end{tabularx}
\label{tab:agent-tools}
\end{table}

\subsection{Worker}
\label{apx:system-worker}

\subsubsection{Logical execution modes}
\sysname supports two execution modes, a \textit{full agent} mode and \textit{LLM-only} mode.
When using the full agent model, a dedicated, isolated execution space is created for the purpose of processing that individual subtask.
We implement these isolated execution spaces using a collection of Docker containers attached to a shared Docker network, but other implementations are possible.
When developing \sysname, we recognized that many subtasks can be realized in the form of a single LLM query with some data attached.
For example, text summarization fits this paradigm.
Therefore, to improve subtask throughput dynamically based on expected subtask complexity, we add an LLM-only execution path.
This path does not create the full execution environment and implements subtask execution by making a single LLM query, formatting the agent request and all data resources inline with the LLM context.
This allows us to process larger sub-batches within a single worker process, which utilizes system resources more efficiently.
The LLM-only execution path has the limitation that the worker agent cannot call tools or access a full computer environment, so it is only suitable for simple subtask types.

\subsubsection{Worker execution environment}
All full agent mode subtasks execute in a Docker-managed execution sandbox.
The sandbox consists of several Docker containers connected over a shared Docker network.
The subtask spec should contain all information (configurations, task details, setup metadata) required to specify which containers to start, how to configure them, and how to connect them.

The top-level component responsible for orchestrating the Docker environments is the ``subtask manager.''
The subtask manager is responsible for reading the subtask spec, starting the necessary containers, and wiring up the network connections between them.
The subtask manager also monitors the health of the containers and handles any necessary cleanup after the subtask execution is complete.
The subtask manager executes directly on the host machine alongside other subtask managers.
When using the ray orchestration layer, the subtask manager implementation serves as the entrypoint for a ray task.

The core container hosts a so-called "leader agent" which is responsible for overseeing the general lifecycle of the subtask.
The leader agent directly interfaces with the subtask inputs and is equipped with tools that correspond with capabilities that originate from the capability registry and which are contained in the subtask spec.
The final output of the subtask is written by the leader agent to a pre-specified output path.
This output will be schema-validated by the subtask manager before propagating the result to the originating task manager executor.

Within the context of a subtask execution fabric, a "helper agent" refers to a long-lived container (concurrent with the leader agent) that exposes a well-defined interface (details of which are included in the capability registry).
These containers are created at subtask startup and persist for the entire duration of the subtask execution.
These containers are therefore allowed to maintain state and execute long-running processes concurrently with the leader agent.

Helper agents are not permitted to communicate with each other, and only have a line of communication with the leader agent.
Helper agents are also not permitted to spawn other helper agents or helper tools.

\subsubsection{Worker agent}

To simplify connection with external data resources and improve resource efficiency, the worker environment does not directly open a connection with the object store backend, but rather connects to a node-level \textit{object store proxy} (details in \Cref{apx:system-cluster}).

\subsection{Executor}
\label{apx:system-executor}

\subsubsection{Automatic retries}
A well-studied challenge faced by distributed computation frameworks is how to seamlessly address transient resource failures such as network partitions \cite{dean2004mapreduce,zaharia2012resilient,moritz2018ray}.
A common solution is to simply retry subtasks that fail due to potentially transient errors.
We adopt this approach at the subtask level: each subtask is attempted up to a maximum of 3 times.
If any attempted execution succeeds, its output is persisted to stable storage and propagated through the batch result to the task manager.
If all execution attempts fail, a logical subtask failure is forwarded instead to the task manager.
This failure resolution is important for providing a clean logical layer to the manager agent.

\subsection{Planning for parallelization}
\label{apx:system-parallel-planning}
A common design pattern in prior research \cite{fourney2024magentic-one}
investigates how to use dedicated planning phases to improve LLM agent performance on long-horizon tasks.
To encourage parallel workflow reasoning, we adapt this design pattern and explicitly model the planning process as a first-class system abstraction.
We realize two forms of planning abstractions, each realized as local manager state maintained by the manager agent.
The first is a local planning structure that encodes high-level parallelization strategy in a flexible, generic format.
The second encodes a formal task contract that self-imposes global state preconditions gating task termination.

\paragraph{Parallel planning structure}
We implement a simple planning abstraction to encourage parallelization-focused planning from the manager agent.
Our parallel plan is designed to support \sysname's goal of automated task decomposition into parallelizable workflows.
Thus, we define the plan structure with explicit consideration of the manager's parallelization objectives.
An \sysname plan consists of two main fields: the \textit{partition strategy} and a general-purpose \textit{list of steps}.
The partition strategy is intended to be populated with explicit data and task partition axes that eventually correspond to batches of subtasks that can be executed in parallel.
The partition strategy is intended to represent a high-level parallelization strategy and be roughly fixed once established.
To enable more granular planning controls, the planning structure also supports a general-purpose list of steps that identify the key procedural table manipulation and batch execution operations that will lead to task completion.
These steps are annotated with agent-provided progress flags, allowing the manager agent to track progress towards individual subgoals.

To clarify the distinguishing conceptual properties of our plan structure, we contrast our planning abstraction with the planning abstraction from Magentic-One \cite{fourney2024magentic-one}.
In addition to facilitating general-purpose planning, Magentic-One's two-tiered planning mechanism is structured primarily around the organization of \textit{facts} anticipated and encountered during agent execution, determining the immediate next agent invocation, detecting unproductive agent loops, and deciding appropriate termination conditions.
This planning abstraction is well-suited for, e.g., knowledge gathering work, but does not facilitate dynamically-discovered parallelized workflows.
In contrast, our planning abstraction explicitly models parallel task decomposition as a first-class consideration.

The manager agent can update the plan structure at any time during its exploration phase using a dedicated tool (see \Cref{apx:task-manager-tools}).

\paragraph{Output contracts}
The manager's plan object primarily functions as a free-form scratch space whose contents are not formalized into any hard execution requirements.
Its main purpose is to refine the manager agent's focus towards achieving the user task as well as to serve as a simple kind of agent memory.
Depending on the task, it may also be possible to derive formal output specifications defining preconditions required for task completion.
We implement this concept in the form of \textit{output contracts}.

Our implementation of output contracts is grounded in \sysname's data table abstraction.
An output contract consists of a set of \textit{table specifications}.
A table specification defines a table that must exist before the task can be considered complete.
Each table specification declares a name for the expected table, as well as (optionally) the column schema and number of rows the table is expected to have.

Agent-generated output contracts are used in a few different ways during system execution.
Most immediately, the task contract (and intermediate progress towards its fulfillment) is formatted into the manager agent's context window.
We hypothesize that conditioning LLM generation on this formatted intermediate progress focuses execution towards higher-utility agent trajectories.
Second, when the manager agent signals task completion, we directly validate the manager table state against the contract in a rule-based manner.
If the contract is unfulfilled, the task manager's control flow is returned to the manager agent to continue the exploration phase.
Although simple, we find that simply encouraging the manager agent to formalize its task objectives as hard execution contracts can prevent premature task termination as well as improve output fidelity against user-specified formatting constraints.

Like the free-form parallelization plan, the manager agent can update the output contract at any point during its exploration phase using a dedicated tool (see \Cref{apx:task-manager-tools}).

\subsection{Subtask delegation}
\label{apx:system-subtask-delegation}
One of the novel challenges that arises from using LLM-based components to specify task decomposition into parallelizable workflows lies in how to simply \textit{specify} that partitioning efficiently.
Within standard (non-agentic) distributed execution engines, decomposition happens at a time scale typically orders of magnitude faster than the time taken 
In particular, it is relatively inexpensive for the programmer to specify thousands, or even millions of subtasks without noticeable overhead compared with the time taken to execute those subtasks.
In contrast, with LLM-based delegation mechanisms, this dynamic is inverted.
Defining a single subtask using a natural language specification can take on the order of seconds, introducing network and text generation overhead.

To resolve this issue, we introduce \textit{subtask templates}.
Subtask templates provide a way for LLM-based components to efficiently specify a large number of subtasks that can be executed in parallel.

\subsection{Data tables}
\label{apx:system-tables}
A core difficulty our architecture must grapple with is how to reason over large-scale data and distributed execution resources.
For the types of tasks we consider, the data can be too large to fit on a single machine (making it impossible to store or examine all intermediate state on a single machine), and even the size of the metadata describing that data can quickly exhaust even frontier LLMs' context windows.
The key logical abstraction we introduce to facilitate agent-based reasoning over large volumes of data is that of a \textit{data table}.

A data table is a logical, read-only, finite sequence of records.
A record is mapping from field names to typed values.
All records in a dataset conform to a common \textit{schema}, which fixes the set of typed fields.
Types determine the admissible domain for field values.

Various operations can be used to produce new datasets from existing ones, forming a \textit{table algebra}.
These operations derive from the standard relational algebra operations like joins, intersections, unions, and filters.

Operators in this algebra are both \textit{pure} (not modifying their inputs) and \textit{deterministic}.
Thus, sequences of nested operations form a declarative specification for how to construct a dataset from a collection of leaf datasets.

All tables in our representation are either \textit{leaf tables} grounded in a metadata-laden stable storage, or are constructed by application of operators in the algebra to one or more dependencies.
Thus, all non-leaf tables are fully specified by their \textit{lineage}.
A lineage (graph) is a directed acyclic graph whose nodes denote operation applications, whose leaves denote base datasets, and whose edges describe data dependencies.
Knowledge of the lineage graph of a dataset is sufficient to reconstruct it from its root dependencies, provided the leaf datasets are available.

Our table abstraction has several key advantages.
First, the table abstraction makes LLM-based reasoning about intermediate task states more tractable.
Table metadata and summaries can be efficiently formatted to fit inside the context window of an LLM, while hiding high-volume data contents.
An LLM-based component integrated with the table subsystem can efficiently specify highly data-parallel transformations, including local table manipulation as well

Second, immutability simplifies coordination with other distributed system components.
In particular, consistency of replicated shared objects is achieved without the need for convoluted consensus protocols.
This property also makes it easier to recompute missing objects using lineage metadata.

We note some explicit restrictions of our abstract data model.
It does not allow for efficient fine-grained memory controls inexpressible with the supported algebraic manipulations.

\subsection{Capability registry}
\label{apx:system-capability-registry}
The \textit{Capability Registry} is a service exposing diverse software functionalities that can be used during subtask execution.
These capabilities enable dynamic, composable functionalities for workers.
We assume the capability registry is implemented as a horizontally-scalable, network-accessible, external service.

Capabilities are categorized based on their local execution model.
\textit{Service} capabilities inhabit long-running execution spaces that persist for the entire duration of its associated subtask's execution environment.
Notably, service capabilities may perform work in parallel with the main subtask driver process.
This property is useful for performing expensive, long-running computations concurrently with a subtask driver process or enabling state persistence.
\textit{Tool} capabilities are stateless interfaces whose execution duration is scoped only for a single invocation.

Entries in the Capability Registry consist of metadata (capability name, unique identifier, natural language description) and implementation location (e.g., a Docker image).
In our application, capabilities are discovered by the task manager during the planning and analysis phases and used to perform specialized operations during subtask execution.

We distinguish our capability registry from existing tool discovery architectures to clarify its contributions.
Our capability registry is substantially different from the Model Context Protocol (MCP) \cite{mcp}, due to MCP's remote execution model.
MCP exposes a function call interface through a standardized network API that facilitates tool discovery and tool invocation.
This differs from our capability registry, which does not perform tool execution but instead points to an implementation of the tool software.

\subsection{Data resources and state management}
\label{apx:system-data}
\sysname assumes access to highly available and durable network-accessed storage resource.
We leverage two storage interfaces: the \textit{state store} and the \textit{object store}.

\paragraph{State store}
The state store maintains all control state for the full set of tasks being processed by the \sysname system.
This includes task submissions, task manager state checkpoints, subtask definitions, subtask execution attempts, subtask outputs, and an append-only event log.

\paragraph{Object store}
The object store is specifically used for storage of large, unstructured data objects.
This interface is particularly well-suited for passing data back and forth between subtask workers and the task manager and can handle text, binary, image, and other data formats.

Objects in the object store are \textit{content-addressed} by a hash of their contents, meaning that URI contents are immutable.
This greatly simplifies data coordination, as updates are only propagated by passing new \textit{handles} to data.
This property also implicitly restricts the ability of agents to use the data plane for communication, since they cannot send messages by modifying the contents pointed at by a given URI.

\subsection{Cluster management}
\label{apx:system-cluster}
In this subsection, we discuss considerations for managing computational and data resources in the cluster computing environment.

\subsubsection{Node resource management}
Ray exposes logical resource controls that constrain the number of co-located subtasks executing on the same Ray node.
These controls are flexible and allow constraints on resource use ranging from (virtual) hardware resources such as CPUs and GPUs to logical resources (for example, limiting port).
We use this mechanism to control the maximum node-level subtask concurrency by configuring each subtask execution to consume 1 logical CPU resource.
Each ray node connected to the cluster is registered with a number of (logical) cores,

\subsubsection{Shared node services}
Each node in \sysname's computing cluster maintains a small set of necessary services to facilitate basic interactions with core \sysname resources.
One of these services is the \textit{object store proxy}.
The object store proxy sits interfaces between subtask execution environments and the true system-wide object store.
Using a proxy service simplifies worker environment setup, as it allows  workers to use a consistent, local network location for accessing data resources in the object store.
Using a proxy service also allows caching of resources, which allows multiple co-located worker processes to access the same data resources at a lower amortized cost.
Crucially, because objects are content-addressed, this cache does not introduce the usual cache staleness concerns.

\section{Extended Experimental Results}
\label{apx:extended-experiments}
In this appendix, we provide detailed results for the experiments discussed in \Cref{sec:evaluation}.

\subsection{Results on \sumtask{}}
\label{apx:full-summarization-results}

\paragraph{Task description}

A description of the dataset used for hierarchical summarization is given in \Cref{tab:summarization-data}.

\begin{table}[!htbp]
    \centering
    \scriptsize
    \setlength{\tabcolsep}{2pt}
    \caption{Description of datasets used for the hierarchical summarization task.}
    \begin{tabular}{@{}c ccccc@{}}
    \toprule
        Corpus & Level 1 & Level 2 & Level 3 & Level 4 & Size (\qty{}{\kilo\byte})  \\
    \midrule
        Romeo and Juliet (R\&J)  & 26 & 5 & - & - & 166 \\
        The Dynasts & 130 & 19 & 3 & - & 942 \\
        Decline and Fall of the Roman Empire (Roman) & 2588 & 296 & 71 & 6 & 10,500 \\
    \bottomrule
    \end{tabular}
    \label{tab:summarization-data}
\end{table}

For hierarchical summarization, we decompose structural scoring into a series of binary checks on the integrity of the output tables generated by the system.
The precise number of conditions scored depends on the particular corpus being processed.
In general, for a given corpus, each layer of the organization hierarchy generates three distinct checks.
First, a ``present'' score checks whether or not the appropriate 
The overall structural score is computed as the mean score of these binary checks.

We apply these same metrics for the baselines used in our evaluation, suitably adapted for the different data representation.
Here, ``presence'' of a table indicates the presence of the appropriate key in the generated JSON object emitted by the system.
Semantic scores are defined identically.

\paragraph{Full results}
We provide detailed results for \sumtask{} in \Cref{tab:hsum-frwc-appendix-results,tab:hsum-strsem-appendix-results,tab:hsum-detail}.

\begin{table*}[!htbp]
\centering
\scriptsize
\setlength{\tabcolsep}{2pt}
\caption{Failure rates and wall-clock runtimes for \sumtask{} across baselines and \sysname{} variants. FR is computed over all selected trials. WC is computed over trials that produced output and reported as mean $\pm$ standard deviation.}
\label{tab:hsum-frwc-appendix-results}
\begin{tabular}{l cc cc cc}
\toprule
 & \multicolumn{2}{c}{R\&J} & \multicolumn{2}{c}{Dynasts} & \multicolumn{2}{c}{Roman} \\
\cmidrule(lr){2-3}
\cmidrule(lr){4-5}
\cmidrule(lr){6-7}
\textbf{Method} & \textbf{FR} & \textbf{WC} & \textbf{FR} & \textbf{WC} & \textbf{FR} & \textbf{WC} \\
\midrule
Direct & 0\% & 19 $\pm$ 1 & 60\% & 76 $\pm$ 0 & 100\% & $\bot$ \\
Magentic-One & 100\% & $\bot$ & 100\% & $\bot$ & 100\% & $\bot$ \\
MegaAgent & 60\% & 472 $\pm$ 44 & 60\% & 248 $\pm$ 63 & 40\% & 579 $\pm$ 189 \\
\midrule
\sysname (\texttt{mini}$\times$\texttt{mini}) & 0\% & 96 $\pm$ 19 & 0\% & 142 $\pm$ 20 & 0\% & 272 $\pm$ 53 \\
\sysname (\texttt{mini}$\times$\texttt{nano}) & 11\% & 85 $\pm$ 9 & 22\% & 142 $\pm$ 29 & 11\% & 281 $\pm$ 45 \\
\sysname (\texttt{5.4}$\times$\texttt{mini}) & 0\% & 168 $\pm$ 17 & 0\% & 235 $\pm$ 27 & 0\% & 372 $\pm$ 46 \\
\sysname (\texttt{5.4}$\times$\texttt{nano}) & 0\% & 174 $\pm$ 21 & 0\% & 234 $\pm$ 30 & 0\% & 390 $\pm$ 43 \\
\bottomrule
\end{tabular}
\end{table*}

\begin{table*}[!htbp]
\centering
\scriptsize
\setlength{\tabcolsep}{2pt}
\caption{Structural and semantic scores for \sumtask{} across baselines and \sysname{} variants. Scores are computed over trials that produced output and reported as mean $\pm$ standard deviation.}
\label{tab:hsum-strsem-appendix-results}
\begin{tabular}{l cc cc cc}
\toprule
 & \multicolumn{2}{c}{R\&J} & \multicolumn{2}{c}{Dynasts} & \multicolumn{2}{c}{Roman} \\
\cmidrule(lr){2-3}
\cmidrule(lr){4-5}
\cmidrule(lr){6-7}
\textbf{Method} & \textbf{Str.} & \textbf{Sem.} & \textbf{Str.} & \textbf{Sem.} & \textbf{Str.} & \textbf{Sem.} \\
\midrule
Direct & 1.000 $\pm$ 0.000 & 0.433 $\pm$ 0.013 & 0.979 $\pm$ 0.021 & 0.210 $\pm$ 0.005 & $\bot$ & $\bot$ \\
Magentic-One & $\bot$ & $\bot$ & $\bot$ & $\bot$ & $\bot$ & $\bot$ \\
MegaAgent & 0.140 $\pm$ 0.140 & 0.043 $\pm$ 0.043 & 0.212 $\pm$ 0.121 & 0.023 $\pm$ 0.017 & 0.160 $\pm$ 0.226 & 0.016 $\pm$ 0.022 \\
\midrule
\sysname (\texttt{mini}$\times$\texttt{mini}) & 0.954 $\pm$ 0.075 & 0.424 $\pm$ 0.136 & 0.971 $\pm$ 0.034 & 0.447 $\pm$ 0.055 & 0.908 $\pm$ 0.133 & 0.230 $\pm$ 0.164 \\
\sysname (\texttt{mini}$\times$\texttt{nano}) & 0.950 $\pm$ 0.037 & 0.426 $\pm$ 0.036 & 0.872 $\pm$ 0.134 & 0.323 $\pm$ 0.165 & 0.914 $\pm$ 0.138 & 0.295 $\pm$ 0.129 \\
\sysname (\texttt{5.4}$\times$\texttt{mini}) & 1.000 $\pm$ 0.001 & 0.528 $\pm$ 0.012 & 0.997 $\pm$ 0.003 & 0.451 $\pm$ 0.025 & 0.983 $\pm$ 0.025 & 0.370 $\pm$ 0.024 \\
\sysname (\texttt{5.4}$\times$\texttt{nano}) & 0.943 $\pm$ 0.033 & 0.439 $\pm$ 0.012 & 0.923 $\pm$ 0.046 & 0.419 $\pm$ 0.018 & 0.872 $\pm$ 0.011 & 0.395 $\pm$ 0.007 \\
\bottomrule
\end{tabular}
\end{table*}

\begin{table}[!htbp]
\centering
\scriptsize
\setlength{\tabcolsep}{2pt}
\caption{\sumtask{} detailed metrics by \sysname{} configuration, corpus, and hierarchy layer. Values are averaged over APWA trials that produced the corresponding layer output.}
\label{tab:hsum-detail}
\begin{tabular}{lllcccccc}
\toprule
 & & & \multicolumn{3}{c}{\textbf{ROUGE}} & \multicolumn{3}{c}{\textbf{Structural}} \\
\cmidrule(lr){4-6}\cmidrule(lr){7-9}
\textbf{Config} & \textbf{Corpus} & \textbf{Layer} & \textbf{R-1} & \textbf{R-2} & \textbf{R-L} & \textbf{Present} & \textbf{URI res.} & \textbf{Word cap} \\
\midrule
  \multirow{12}{*}{\sysname (\texttt{mini}$\times$\texttt{mini})} & \multirow{3}{*}{R\&J} & Scene & 0.349 & 0.120 & 0.226 & 1.000 & 0.857 & 0.604 \\
   &  & Act & 0.411 & 0.127 & 0.237 & 1.000 & 0.857 & 0.857 \\
   &  & Full & 0.514 & 0.144 & 0.319 & 1.000 & 1.000 & 1.000 \\
  \cmidrule{2-9}
   & \multirow{4}{*}{Dyn.} & Scene & 0.409 & 0.136 & 0.245 & 1.000 & 0.857 & 0.714 \\
   &  & Act & 0.469 & 0.123 & 0.242 & 1.000 & 1.000 & 0.939 \\
   &  & Part & 0.427 & 0.111 & 0.225 & 1.000 & 1.000 & 0.943 \\
   &  & Full & 0.484 & 0.115 & 0.219 & 1.000 & 1.000 & 0.857 \\
  \cmidrule{2-9}
   & \multirow{5}{*}{Roman} & Paragraph & 0.355 & 0.106 & 0.222 & 1.000 & 0.857 & 0.857 \\
   &  & Part & 0.224 & 0.033 & 0.101 & 1.000 & 0.714 & 0.705 \\
   &  & Chapter & 0.206 & 0.026 & 0.090 & 1.000 & 0.714 & 0.692 \\
   &  & Volume & 0.196 & 0.022 & 0.081 & 1.000 & 0.714 & 0.690 \\
   &  & Full & 0.170 & 0.015 & 0.075 & 1.000 & 0.714 & 0.571 \\
\midrule
  \multirow{12}{*}{\sysname (\texttt{mini}$\times$\texttt{nano})} & \multirow{3}{*}{R\&J} & Scene & 0.397 & 0.107 & 0.227 & 1.000 & 1.000 & 0.821 \\
   &  & Act & 0.394 & 0.096 & 0.210 & 1.000 & 1.000 & 0.556 \\
   &  & Full & 0.434 & 0.097 & 0.228 & 0.889 & 0.889 & 0.667 \\
  \cmidrule{2-9}
   & \multirow{4}{*}{Dyn.} & Scene & 0.444 & 0.137 & 0.255 & 1.000 & 1.000 & 0.619 \\
   &  & Act & 0.342 & 0.072 & 0.163 & 1.000 & 0.889 & 0.291 \\
   &  & Part & 0.253 & 0.052 & 0.121 & 0.889 & 0.778 & 0.578 \\
   &  & Full & 0.218 & 0.037 & 0.098 & 0.667 & 0.667 & 0.222 \\
  \cmidrule{2-9}
   & \multirow{5}{*}{Roman} & Paragraph & 0.472 & 0.161 & 0.307 & 1.000 & 1.000 & 0.882 \\
   &  & Part & 0.311 & 0.042 & 0.135 & 1.000 & 0.999 & 0.634 \\
   &  & Chapter & 0.233 & 0.023 & 0.099 & 0.889 & 0.889 & 0.642 \\
   &  & Volume & 0.201 & 0.017 & 0.081 & 0.778 & 0.778 & 0.648 \\
   &  & Full & 0.198 & 0.013 & 0.080 & 0.778 & 0.778 & 0.667 \\
\midrule
  \multirow{12}{*}{\sysname (\texttt{5.4}$\times$\texttt{mini})} & \multirow{3}{*}{R\&J} & Scene & 0.530 & 0.184 & 0.341 & 1.000 & 1.000 & 0.996 \\
   &  & Act & 0.475 & 0.149 & 0.263 & 1.000 & 1.000 & 1.000 \\
   &  & Full & 0.578 & 0.180 & 0.363 & 1.000 & 1.000 & 1.000 \\
  \cmidrule{2-9}
   & \multirow{4}{*}{Dyn.} & Scene & 0.524 & 0.174 & 0.311 & 1.000 & 1.000 & 1.000 \\
   &  & Act & 0.458 & 0.119 & 0.236 & 1.000 & 1.000 & 0.958 \\
   &  & Part & 0.415 & 0.104 & 0.222 & 1.000 & 1.000 & 0.978 \\
   &  & Full & 0.409 & 0.090 & 0.202 & 1.000 & 1.000 & 1.000 \\
  \cmidrule{2-9}
   & \multirow{5}{*}{Roman} & Paragraph & 0.471 & 0.136 & 0.288 & 1.000 & 1.000 & 1.000 \\
   &  & Part & 0.386 & 0.056 & 0.174 & 1.000 & 1.000 & 0.962 \\
   &  & Chapter & 0.359 & 0.044 & 0.155 & 1.000 & 1.000 & 0.925 \\
   &  & Volume & 0.312 & 0.032 & 0.132 & 1.000 & 1.000 & 0.926 \\
   &  & Full & 0.324 & 0.022 & 0.136 & 1.000 & 1.000 & 0.667 \\
\midrule
  \multirow{12}{*}{\sysname (\texttt{5.4}$\times$\texttt{nano})} & \multirow{3}{*}{R\&J} & Scene & 0.436 & 0.120 & 0.253 & 1.000 & 1.000 & 0.949 \\
   &  & Act & 0.390 & 0.096 & 0.211 & 1.000 & 1.000 & 0.587 \\
   &  & Full & 0.491 & 0.111 & 0.249 & 1.000 & 1.000 & 0.444 \\
  \cmidrule{2-9}
   & \multirow{4}{*}{Dyn.} & Scene & 0.500 & 0.155 & 0.286 & 1.000 & 1.000 & 0.682 \\
   &  & Act & 0.427 & 0.090 & 0.202 & 1.000 & 1.000 & 0.323 \\
   &  & Part & 0.375 & 0.077 & 0.184 & 1.000 & 1.000 & 0.600 \\
   &  & Full & 0.374 & 0.065 & 0.164 & 1.000 & 1.000 & 0.556 \\
  \cmidrule{2-9}
   & \multirow{5}{*}{Roman} & Paragraph & 0.508 & 0.159 & 0.316 & 1.000 & 1.000 & 0.750 \\
   &  & Part & 0.404 & 0.055 & 0.170 & 1.000 & 1.000 & 0.119 \\
   &  & Chapter & 0.372 & 0.040 & 0.148 & 1.000 & 1.000 & 0.102 \\
   &  & Volume & 0.346 & 0.031 & 0.136 & 1.000 & 1.000 & 0.204 \\
   &  & Full & 0.347 & 0.028 & 0.135 & 1.000 & 1.000 & 0.000 \\
\bottomrule
\end{tabular}
\end{table}

\subsection{Results on \piitask{}}
\label{apx:full-pii-results}

We report the full \piitask{} results in \Cref{tab:pii-frwc-appendix-results,tab:pii-strsem-appendix-results,tab:pii-detail}.

\begin{table*}[!htbp]
\centering
\scriptsize
\setlength{\tabcolsep}{2pt}
\caption{Failure rates and wall-clock runtimes for \piitask{} across baselines and \sysname{} variants. FR is computed over all selected trials. WC is computed over trials that produced output and reported as mean $\pm$ standard deviation.}
\label{tab:pii-frwc-appendix-results}
\begin{tabular}{l cc cc cc}
\toprule
 & \multicolumn{2}{c}{PII-64} & \multicolumn{2}{c}{PII-512} & \multicolumn{2}{c}{PII-4096} \\
\cmidrule(lr){2-3}
\cmidrule(lr){4-5}
\cmidrule(lr){6-7}
\textbf{Method} & \textbf{FR} & \textbf{WC} & \textbf{FR} & \textbf{WC} & \textbf{FR} & \textbf{WC} \\
\midrule
Direct & 0\% & 37 $\pm$ 4 & 0\% & 41 $\pm$ 29 & 100\% & $\bot$ \\
Magentic-One & 100\% & $\bot$ & 100\% & $\bot$ & 80\% & 91 $\pm$ 0 \\
MegaAgent & 60\% & 390 $\pm$ 266 & 80\% & 25 & 70\% & 372 $\pm$ 488 \\
\midrule
\sysname (\texttt{mini}$\times$\texttt{mini}) & 0\% & 35 $\pm$ 4 & 0\% & 67 $\pm$ 52 & 0\% & 221 $\pm$ 119 \\
\bottomrule
\end{tabular}
\end{table*}

\begin{table*}[t]
\centering
\scriptsize
\setlength{\tabcolsep}{2pt}
\caption{Structural and semantic scores for \piitask{} across baselines and \sysname{} variants. Scores are computed over trials that produced output and reported as mean $\pm$ standard deviation.}
\label{tab:pii-strsem-appendix-results}
\begin{tabular}{l cc cc cc}
\toprule
 & \multicolumn{2}{c}{PII-64} & \multicolumn{2}{c}{PII-512} & \multicolumn{2}{c}{PII-4096} \\
\cmidrule(lr){2-3}
\cmidrule(lr){4-5}
\cmidrule(lr){6-7}
\textbf{Method} & \textbf{Str.} & \textbf{Sem.} & \textbf{Str.} & \textbf{Sem.} & \textbf{Str.} & \textbf{Sem.} \\
\midrule
Direct & 1.000 $\pm$ 0.000 & 0.775 $\pm$ 0.046 & 0.750 $\pm$ 0.000 & 0.162 $\pm$ 0.096 & $\bot$ & $\bot$ \\
Magentic-One & $\bot$ & $\bot$ & $\bot$ & $\bot$ & 1.000 $\pm$ 0.000 & 0.179 $\pm$ 0.076 \\
MegaAgent & 0.375 $\pm$ 0.375 & 0.000 $\pm$ 0.000 & 0.000 & 0.000 & 0.250 $\pm$ 0.354 & 0.000 $\pm$ 0.000 \\
\midrule
\sysname (\texttt{mini}$\times$\texttt{mini}) & 1.000 $\pm$ 0.000 & 0.759 $\pm$ 0.064 & 1.000 $\pm$ 0.000 & 0.772 $\pm$ 0.040 & 0.900 $\pm$ 0.300 & 0.544 $\pm$ 0.356 \\
\bottomrule
\end{tabular}
\end{table*}

\begin{table}[!htbp]
\centering
\scriptsize
\setlength{\tabcolsep}{3pt}
\caption{\piitask{} detailed structural and semantic metrics by \sysname{} configuration and document count. Values are averaged over APWA trials that produced output.}
\label{tab:pii-detail}
\begin{tabular}{llcccccc}
\toprule
\textbf{Config} & \textbf{Size} & \textbf{Struct.} & \textbf{PII Prec.} & \textbf{PII Recall} & \textbf{PII F1} & \textbf{Content Sim.} & \textbf{Exact Match} \\
\midrule
  \multirow{3}{*}{\sysname (\texttt{mini}$\times$\texttt{mini})} & PII-64 & 1.000 & 0.833 & 0.758 & 0.793 & 0.894 & 0.347 \\
   & PII-512 & 1.000 & 0.840 & 0.757 & 0.796 & 0.893 & 0.350 \\
   & PII-4096 & 0.900 & 0.593 & 0.528 & 0.559 & 0.694 & 0.244 \\
\bottomrule
\end{tabular}
\end{table}

\subsection{Results on \schematask{}}
\label{apx:full-schema-results}

We report full results for \schematask{} experiments in \Cref{tab:schema-frwc-appendix-results,tab:schema-strsem-appendix-results,tab:schema-detail}.

\begin{table*}[!htbp]
\centering
\scriptsize
\setlength{\tabcolsep}{1.5pt}
\caption{Failure rates and wall-clock runtimes for \schematask{} across baselines and \sysname{} variants. FR is computed over all selected trials. WC is computed over trials that produced output and reported as mean $\pm$ standard deviation.}
\label{tab:schema-frwc-appendix-results}
\begin{tabular}{l cc cc cc cc}
\toprule
 & \multicolumn{2}{c}{\chemistry{}} & \multicolumn{2}{c}{\mltables{}} & \multicolumn{2}{c}{\discomat{}} & \multicolumn{2}{c}{\swde{}} \\
\cmidrule(lr){2-3}
\cmidrule(lr){4-5}
\cmidrule(lr){6-7}
\cmidrule(lr){8-9}
\textbf{Method} & \textbf{FR} & \textbf{WC} & \textbf{FR} & \textbf{WC} & \textbf{FR} & \textbf{WC} & \textbf{FR} & \textbf{WC} \\
\midrule
Direct & 0\% & 350 $\pm$ 44 & 100\% & $\bot$ & 100\% & $\bot$ & 100\% & $\bot$ \\
Magentic-One & 80\% & 95 & 60\% & 79 $\pm$ 52 & 100\% & $\bot$ & 60\% & 88 $\pm$ 47 \\
MegaAgent & 60\% & 141 $\pm$ 53 & 100\% & $\bot$ & 40\% & 278 $\pm$ 202 & 80\% & 1222 \\
\midrule
\sysname (\texttt{mini}$\times$\texttt{mini}) & 10\% & 62 $\pm$ 17 & 10\% & 137 $\pm$ 25 & 10\% & 103 $\pm$ 22 & 10\% & 104 $\pm$ 39 \\
\sysname (\texttt{mini}$\times$\texttt{nano}) & 10\% & 75 $\pm$ 14 & 10\% & 191 $\pm$ 57 & 0\% & 101 $\pm$ 49 & 0\% & 95 $\pm$ 20 \\
\sysname (\texttt{5.4}$\times$\texttt{mini}) & 0\% & 95 $\pm$ 10 & 0\% & 183 $\pm$ 22 & 0\% & 147 $\pm$ 13 & 10\% & 125 $\pm$ 11 \\
\sysname (\texttt{5.4}$\times$\texttt{nano}) & 10\% & 115 $\pm$ 8 & 10\% & 231 $\pm$ 37 & 10\% & 187 $\pm$ 27 & 10\% & 128 $\pm$ 17 \\
\bottomrule
\end{tabular}
\end{table*}

\begin{table*}[!htbp]
\centering
\scriptsize
\setlength{\tabcolsep}{1.5pt}
\caption{Structural and semantic scores for \schematask{} across baselines and \sysname{} variants. Scores are computed over trials that produced output and reported as mean $\pm$ standard deviation.}
\label{tab:schema-strsem-appendix-results}
\begin{tabular}{l cc cc cc cc}
\toprule
 & \multicolumn{2}{c}{\chemistry{}} & \multicolumn{2}{c}{\mltables{}} & \multicolumn{2}{c}{\discomat{}} & \multicolumn{2}{c}{\swde{}} \\
\cmidrule(lr){2-3}
\cmidrule(lr){4-5}
\cmidrule(lr){6-7}
\cmidrule(lr){8-9}
\textbf{Method} & \textbf{Str.} & \textbf{Sem.} & \textbf{Str.} & \textbf{Sem.} & \textbf{Str.} & \textbf{Sem.} & \textbf{Str.} & \textbf{Sem.} \\
\midrule
Direct & 0.667 $\pm$ 0.000 & 0.534 $\pm$ 0.091 & $\bot$ & $\bot$ & $\bot$ & $\bot$ & $\bot$ & $\bot$ \\
Magentic-One & 1.000 & 0.459 & 1.000 $\pm$ 0.000 & 0.054 $\pm$ 0.003 & $\bot$ & $\bot$ & 1.000 $\pm$ 0.000 & 0.259 $\pm$ 0.084 \\
MegaAgent & 0.333 $\pm$ 0.333 & 0.000 $\pm$ 0.000 & $\bot$ & $\bot$ & 0.222 $\pm$ 0.314 & 0.000 $\pm$ 0.000 & 0.667 & 0.000 \\
\midrule
\sysname (\texttt{mini}$\times$\texttt{mini}) & 1.000 $\pm$ 0.000 & 0.705 $\pm$ 0.251 & 1.000 $\pm$ 0.000 & 0.527 $\pm$ 0.115 & 0.963 $\pm$ 0.105 & 0.661 $\pm$ 0.236 & 1.000 $\pm$ 0.000 & 0.856 $\pm$ 0.217 \\
\sysname (\texttt{mini}$\times$\texttt{nano}) & 1.000 $\pm$ 0.000 & 0.583 $\pm$ 0.182 & 1.000 $\pm$ 0.000 & 0.465 $\pm$ 0.133 & 0.722 $\pm$ 0.124 & 0.649 $\pm$ 0.044 & 1.000 $\pm$ 0.000 & 0.812 $\pm$ 0.251 \\
\sysname (\texttt{5.4}$\times$\texttt{mini}) & 1.000 $\pm$ 0.000 & 0.794 $\pm$ 0.027 & 1.000 $\pm$ 0.000 & 0.566 $\pm$ 0.025 & 1.000 $\pm$ 0.000 & 0.801 $\pm$ 0.020 & 1.000 $\pm$ 0.000 & 0.942 $\pm$ 0.001 \\
\sysname (\texttt{5.4}$\times$\texttt{nano}) & 1.000 $\pm$ 0.000 & 0.763 $\pm$ 0.033 & 1.000 $\pm$ 0.000 & 0.549 $\pm$ 0.030 & 0.741 $\pm$ 0.139 & 0.703 $\pm$ 0.048 & 1.000 $\pm$ 0.000 & 0.924 $\pm$ 0.006 \\
\bottomrule
\end{tabular}
\end{table*}

\begin{table}[!htbp]
\centering
\scriptsize
\setlength{\tabcolsep}{3pt}
\caption{\schematask{} detailed structural and classification metrics by \sysname{} configuration and schema extraction suite. Values are averaged over APWA trials that produced output.}
\label{tab:schema-detail}
\begin{tabular}{llcrrrrrr}
\toprule
 & & & \multicolumn{3}{c}{\textbf{Micro}} & \multicolumn{3}{c}{\textbf{Macro}} \\
\cmidrule(lr){4-6}\cmidrule(lr){7-9}
\textbf{Config} & \textbf{Suite} & \textbf{Struct.} & \textbf{P} & \textbf{R} & \textbf{F1} & \textbf{P} & \textbf{R} & \textbf{F1} \\
\midrule
  \multirow{4}{*}{\sysname (\texttt{mini}$\times$\texttt{mini})} & \chemistry{} & 1.000 & 0.699 & 0.791 & 0.742 & 0.690 & 0.759 & 0.705 \\
   & \mltables{} & 1.000 & 0.671 & 0.546 & 0.583 & 0.588 & 0.519 & 0.527 \\
   & \discomat{} & 0.963 & 0.663 & 0.660 & 0.661 & \tiny{N/A} & \tiny{N/A} & \tiny{N/A} \\
   & \swde{} & 1.000 & \tiny{N/A} & \tiny{N/A} & \tiny{N/A} & \tiny{N/A} & \tiny{N/A} & 0.856 \\
\midrule
  \multirow{4}{*}{\sysname (\texttt{mini}$\times$\texttt{nano})} & \chemistry{} & 1.000 & 0.693 & 0.636 & 0.655 & 0.589 & 0.618 & 0.583 \\
   & \mltables{} & 1.000 & 0.586 & 0.449 & 0.488 & 0.570 & 0.435 & 0.465 \\
   & \discomat{} & 0.722 & 0.672 & 0.629 & 0.649 & \tiny{N/A} & \tiny{N/A} & \tiny{N/A} \\
   & \swde{} & 1.000 & \tiny{N/A} & \tiny{N/A} & \tiny{N/A} & \tiny{N/A} & \tiny{N/A} & 0.812 \\
\midrule
  \multirow{4}{*}{\sysname (\texttt{5.4}$\times$\texttt{mini})} & \chemistry{} & 1.000 & 0.773 & 0.868 & 0.817 & 0.786 & 0.841 & 0.794 \\
   & \mltables{} & 1.000 & 0.670 & 0.626 & 0.646 & 0.578 & 0.569 & 0.566 \\
   & \discomat{} & 1.000 & 0.788 & 0.814 & 0.801 & \tiny{N/A} & \tiny{N/A} & \tiny{N/A} \\
   & \swde{} & 0.900 & \tiny{N/A} & \tiny{N/A} & \tiny{N/A} & \tiny{N/A} & \tiny{N/A} & 0.848 \\
\midrule
  \multirow{4}{*}{\sysname (\texttt{5.4}$\times$\texttt{nano})} & \chemistry{} & 1.000 & 0.766 & 0.810 & 0.786 & 0.760 & 0.819 & 0.763 \\
   & \mltables{} & 1.000 & 0.653 & 0.557 & 0.596 & 0.613 & 0.527 & 0.549 \\
   & \discomat{} & 0.741 & 0.691 & 0.716 & 0.703 & \tiny{N/A} & \tiny{N/A} & \tiny{N/A} \\
   & \swde{} & 1.000 & \tiny{N/A} & \tiny{N/A} & \tiny{N/A} & \tiny{N/A} & \tiny{N/A} & 0.924 \\
\bottomrule
\end{tabular}
\end{table}

\subsection{Results on \websurfer}
\label{apx:full-websurfer-results}

\paragraph{Task description}
The goal of the benchmark is to compose a report (100-200 words) on a manually curated set of topics drawn from diverse domains. It is designed to show that APWA is not only capable of dynamically identifying the capability presets required for a task but is also able to utilize the corresponding agent from the registry at scale. Topics are divided into small (10), moderate (20) and large (100) splits. The benchmark will require the need of an agent with web browsing capabilities for which we use the Websurfer agent from Magentic-One. 

We score APWA along two dimensions: utility (whether the system produces a correct output)  and cost (wall-clock runtime). We decompose utility into two submetrics, structural (whether the outputs are correctly formatted) and semantic (whether the outputs are correct). The structural score measures whether APWA returns the expected number of (topic, report) pairs. The semantic score evaluates each pair on 2 attributes - whether the report addresses the requested topic and whether the content is substantive, factual information rather than filler or off-topic. We use LLM-as-a-judge to assess these attributes, a pair receives a score of 1.0 if both attributes hold, 0.5 if only one holds and 0.0 otherwise. The final semantic utility is a mean over all pairs.

\paragraph{Full results}
On the small split(10 topics), APWA completes the task in 2 min 23 s  while maintaining a perfect score on both structural and semantic utility, successfully producing 10 relevant and factual reports. On the moderate split (20 topics), runtime increases to 2 min 37 s but APWA achieves a perfect structural score while yielding a semantic score of 0.975, with 19 out of 20 topics having a relevant and factual report and 1 deemed as relevant but not substantive.  On the large split (100 topics), runtime grows to 10 min; APWA again achieves a perfect structural score and a semantic utility of 0.995, with 99 of 100 topics yielding a relevant and factual report and 1 deemed relevant but lacking substantive content. Notably, the runtime scaling is sublinear, increasing the workload from 10 to 100 topics (a 10$\times$ increase) results in only a 4.2$\times$ increase in the runtime. This suggests that the execution overhead is progressively amortized across larger batches, indicating favorable scalability for the task.

\end{document}